\documentclass[11pt]{article}

\usepackage[final,nohyperref]{acl}

\usepackage{latexsym}
\usepackage[T2A,T1]{fontenc}
\usepackage[utf8]{inputenc}
\usepackage{times}
\usepackage{inconsolata}
\newcommand{\tj}[1]{{\fontencoding{T2A}\fontfamily{cmr}\selectfont #1}}
\DeclareUnicodeCharacter{04E2}{\={\CYRI}}
\DeclareUnicodeCharacter{04E3}{\={\cyri}}
\DeclareUnicodeCharacter{04EE}{\={\CYRU}}
\DeclareUnicodeCharacter{04EF}{\={\cyru}}
\usepackage{microtype}
\usepackage{graphicx}
\usepackage{booktabs}
\usepackage{enumitem}
\setlist{nosep,leftmargin=*}

\newcommand{\BaseLawRten}{49.3}

\newcommand{\BaseTaxRten}{45.9}

\newcommand{\BaseTaxHi}{55.0}

\newcommand{\BaseWikiRten}{89.6}

\newcommand{\BaseBelTgRone}{51.7}

\newcommand{\BaseBelKkRten}{76.3}
\newcommand{\BaseBelKkRone}{60.8}

\newcommand{\BaseBelKkLo}{73.6}
\newcommand{\BaseBelKkHi}{79.3}

\newcommand{\BaseBelUzRone}{59.7}

\newcommand{\BaseBelKyRone}{59.8}

\newcommand{\BaseWikiKkRten}{93.6}
\newcommand{\BaseWikiKkRone}{84.2}

\newcommand{\BaseWikiKkLo}{91.4}
\newcommand{\BaseWikiKkHi}{95.8}

\newcommand{\BaseRuRten}{98.6}
\newcommand{\BaseRuRone}{91.2}

\newcommand{\BaseEnRten}{99.2}
\newcommand{\BaseEnRone}{93.2}

\newcommand{\TgLawRten}{92.4}

\newcommand{\TgBelTgRone}{76.6}

\newcommand{\TgBelKkRten}{82.0}
\newcommand{\TgBelKkRone}{64.4}

\newcommand{\TgBelKkLo}{79.6}
\newcommand{\TgBelKkHi}{84.7}

\newcommand{\TgBelUzRone}{68.9}

\newcommand{\TgBelKyRone}{65.7}

\newcommand{\TgWikiKkRten}{96.6}
\newcommand{\TgWikiKkRone}{88.8}

\newcommand{\TgWikiKkLo}{95.0}
\newcommand{\TgWikiKkHi}{98.0}

\newcommand{\KkLawRten}{53.7}

\newcommand{\KkTaxRten}{66.1}

\newcommand{\KkTaxLo}{56.9}

\newcommand{\KkWikiRten}{90.8}

\newcommand{\KkBelTgRone}{50.4}

\newcommand{\KkBelKkRten}{91.4}
\newcommand{\KkBelKkRone}{78.0}

\newcommand{\KkBelKkLo}{89.6}
\newcommand{\KkBelKkHi}{93.3}

\newcommand{\KkBelUzRone}{67.0}

\newcommand{\KkBelKyRone}{66.4}

\newcommand{\KkWikiKkRten}{98.4}
\newcommand{\KkWikiKkRone}{91.4}

\newcommand{\KkWikiKkLo}{97.2}
\newcommand{\KkWikiKkHi}{99.4}

\newcommand{\KkRuRten}{98.1}
\newcommand{\KkRuRone}{89.6}

\newcommand{\KkEnRten}{98.4}
\newcommand{\KkEnRone}{92.3}

\newcommand{\CaBelTgRone}{77.7}

\newcommand{\CaBelKkRten}{89.8}
\newcommand{\CaBelKkRone}{75.2}

\newcommand{\CaBelKkLo}{87.9}
\newcommand{\CaBelKkHi}{91.8}

\newcommand{\CaBelUzRone}{75.8}

\newcommand{\CaBelKyRone}{75.4}

\newcommand{\CaWikiKkRten}{98.2}
\newcommand{\CaWikiKkRone}{92.2}

\newcommand{\CaWikiKkLo}{97.0}
\newcommand{\CaWikiKkHi}{99.2}

\newcommand{\EfiveBelKkRten}{95.9}
\newcommand{\EfiveBelKkRone}{82.4}

\newcommand{\EfiveBelKkLo}{94.6}
\newcommand{\EfiveBelKkHi}{97.2}

\newcommand{\EfiveWikiKkRten}{99.2}
\newcommand{\EfiveWikiKkRone}{93.6}

\newcommand{\EfiveWikiKkLo}{98.4}
\newcommand{\EfiveWikiKkHi}{99.8}

\newcommand{\BgeBelKkRten}{95.9}
\newcommand{\BgeBelKkRone}{84.6}

\newcommand{\BgeBelKkLo}{94.6}
\newcommand{\BgeBelKkHi}{97.2}

\newcommand{\EfTgBelKkRten}{96.7}
\newcommand{\EfTgBelKkRone}{85.4}

\newcommand{\EfTgBelKkLo}{95.6}
\newcommand{\EfTgBelKkHi}{97.9}

\newcommand{\EfTgWikiKkRten}{99.4}
\newcommand{\EfTgWikiKkRone}{92.8}

\newcommand{\EfTgWikiKkLo}{98.6}
\newcommand{\EfTgWikiKkHi}{100.0}

\newcommand{\KkVsBaseLawRten}{+4.4}
\newcommand{\KkVsBaseLawRtenAbs}{4.4}

\newcommand{\KkVsBaseLawOverlap}{overlap}

\newcommand{\KkVsBaseTaxRten}{+20.2}
\newcommand{\KkVsBaseTaxRtenAbs}{20.2}

\newcommand{\KkVsBaseTaxOverlap}{do not overlap}

\newcommand{\KkVsBaseWikiRten}{+1.2}

\newcommand{\KkVsBgeBelKkOverlap}{do not overlap}

\newcommand{\KkVsCaBelKkRoneAbs}{2.8}

\newcommand{\KkVsBaseBelKkRtenAbs}{15.1}

\newcommand{\KkVsBaseBelKkRoneAbs}{17.2}
\newcommand{\KkVsBaseBelKkOverlap}{do not overlap}

\newcommand{\TgVsBaseBelKkRoneAbs}{3.6}

\newcommand{\BgeVsBaseBelKkRoneAbs}{23.8}

\newcommand{\KkVsBaseBelUzRoneAbs}{7.3}
\newcommand{\KkVsBaseBelUzOverlap}{do not overlap}

\newcommand{\TgVsBaseBelUzRoneAbs}{9.2}

\newcommand{\KkVsBaseBelKyRoneAbs}{6.6}
\newcommand{\KkVsBaseBelKyOverlap}{do not overlap}

\newcommand{\TgVsBaseBelKyRoneAbs}{5.9}

\newcommand{\KkVsCaWikiKkRoneAbs}{0.8}

\newcommand{\KkVsBaseRuRten}{-0.5}

\newcommand{\KkVsBaseRuRone}{-1.6}

\newcommand{\KkVsBaseEnRten}{-0.8}

\newcommand{\KkVsBaseEnRone}{-0.9}

\newcommand{\MtebBaseBelKkRone}{60.7}

\newcommand{\MtebEfiveBelKkRone}{82.9}

\newcommand{\MtebBgeBelKkRone}{84.0}

\newcommand{\MtebMaxDiff}{2.4}

\newcommand{\FertKk}{3.19}

\newcommand{\FertRu}{1.87}

\newcommand{\NLawQ}{1,000}
\newcommand{\NLawCorpus}{72,920}
\newcommand{\NTaxQ}{109}
\newcommand{\NTaxCorpus}{23,282}

\newcommand{\NBelKkQ}{900}
\newcommand{\NBelKkCorpus}{488}

\newcommand{\NWikiKkQ}{500}
\newcommand{\NWikiKkCorpus}{30,000}

\newcommand{\NPoolKk}{166,263}
\newcommand{\NGroundedKk}{251,778}
\newcommand{\NDropHeldKk}{6,880}

\newcommand{\NDropTestKk}{69,404}

\newcommand{\NSetKkTotal}{208,511}
\newcommand{\NSetKkLang}{166,263}

\newcommand{\StepsKk}{410}

\newcommand{\TrainLr}{$2\times10^{-5}$}

\newcommand{\NLayers}{24}
\newcommand{\Vocab}{262,144}
\newcommand{\EmbDim}{768}
\newcommand{\QueryPrompt}{task: search result \textbar{} query: }
\newcommand{\DocPrompt}{title: none \textbar{} text: }
\newcommand{\KkResumeSame}{6}
\newcommand{\KkLossFirst}{0.58}
\newcommand{\KkLossLastMean}{0.29}
\newcommand{\MiniBatchKk}{8}
\newcommand{\NKkNeg}{153,906}
\newcommand{\NKkNoNeg}{12,357}
\newcommand{\PctKkWithNeg}{93}
\newcommand{\PctKkReplay}{20}
\newcommand{\NKkReplayEn}{24,993}
\newcommand{\NKkReplayRu}{17,255}
\newcommand{\KkSetsMatch}{identical}
\newcommand{\NDropNoPassageKk}{9,218}
\newcommand{\NDropDupKk}{13}
\newcommand{\NUniqPosKk}{93,914}
\newcommand{\KkBgeGapBelKk}{6.6}
\newcommand{\KkEfiveGapBelKk}{4.4}
\newcommand{\KkStepsFromParts}{410}
\newcommand{\KkPlainSteps}{407}
\newcommand{\NDropGuardKk}{0}
\newcommand{\KkPassPct}{65}
\newcommand{\KkHalfMin}{1.9}
\newcommand{\KkHalfMax}{2.9}
\newcommand{\MtebCommit}{62d67f4a05}
\newcommand{\NMtebModels}{194}
\newcommand{\NMtebAbove}{33}
\newcommand{\NMtebAboveSmall}{9}
\newcommand{\NMtebAboveNoSize}{10}
\newcommand{\NMtebBelow}{161}
\newcommand{\GemmaParamsM}{308}
\newcommand{\MtebTopName}{Mira190/Euler-Legal-Embedding-V1}
\newcommand{\MtebTopRone}{92.2}
\newcommand{\KkTopGap}{14.2}
\newcommand{\SmallestAboveName}{lightonai/mSlideOn}
\newcommand{\SmallestAboveParamsM}{339}
\newcommand{\SmallestAboveRone}{80.7}
\newcommand{\SmallCutB}{0.6}
\newcommand{\OpenAiLargeRone}{47.1}

\newcommand{\ColabMaxAnchorDiff}{0.8}
\newcommand{\ColabGpu}{T4}
\newcommand{\NKkSpecific}{6}
\newcommand{\NKkSpecificBelow}{4}
\newcommand{\ColabOursRone}{77.3}
\newcommand{\ColabOursRten}{91.4}

\newcommand{\KazEmbRone}{86.0}

\newcommand{\KazEmbParamsM}{1,695}
\newcommand{\DarmmMultiRone}{81.0}

\newcommand{\DarmmMultiParamsM}{568}
\newcommand{\KkEfiveRagRone}{72.6}
\newcommand{\KkEfiveRagRten}{91.6}

\newcommand{\KazembedVfiveRone}{69.0}

\newcommand{\DarmmKkVtwoRone}{64.3}

\newcommand{\AlphaTrimRone}{57.8}

\newcommand{\ColabBgeRone}{83.8}

\newcommand{\OursVsKkEfiveRagOverlap}{overlap}
\newcommand{\OursVsKkEfiveRagRoneAbs}{4.7}
\newcommand{\OursVsKazembedVfiveOverlap}{overlap}

\newcommand{\OursVsKazEmbOverlap}{do not overlap}

\newcommand{\ModelName}{Saidzoda-Kazakh-EmbeddingGemma-300M}

\title{\ModelName:\\How Far One Epoch of Generated Questions Takes a Small Kazakh Embedder}

\author{Tohir Saidzoda \and Madina Abdusalomzoda \and Abubakr Abdullozoda \\
  \textbf{Amin Jalilov} \and \textbf{Faroiz Bobokhon} \and \textbf{Aibek Shokanuly} \\
  Saidzoda AI Lab, Saidzoda Engineering Company, Dushanbe, Tajikistan \\
  \texttt{tohir@saidzoda.com}}

\usepackage{fancyhdr}
\fancypagestyle{plain}{\fancyhf{}\fancyhead[L]{\fontsize{20}{24}\selectfont\textbf{Saidzoda AI Lab}}\fancyfoot[C]{\thepage}}
\begin{document}
\maketitle
\thispagestyle{plain}

\begin{abstract}
EmbeddingGemma-300M is a small open embedder, and used as released it finds the right passage first for \BaseBelKkRone\% of the Kazakh Belebele retrieval questions.
We present \ModelName, the same model fully fine-tuned for one epoch on \NSetKkTotal{} pairs: \NSetKkLang{} Kazakh questions that our own language model wrote about source passages from encyclopedias, Wikipedia, news and books, each checked by a second model, plus English and Russian replay.
On Kazakh Belebele, recall@1 rises from \BaseBelKkRone\% to \KkBelKkRone\% and recall@10 from \BaseBelKkRten\% to \KkBelKkRten\%.
Our model scores above \NMtebBelow{} of the \NMtebModels{} models with a published MTEB score on the same test, among them OpenAI text-embedding-3-large (\OpenAiLargeRone\%).
The other \NMtebAbove{} score higher: the ten best are large multi-billion-parameter or closed models, reaching up to \MtebTopRone\%, and at least \NMtebAboveSmall{} are open models of up to \SmallCutB{}B parameters, the smallest with \SmallestAboveParamsM{}M.
We also scored \NKkSpecific{} Kazakh-specific embedders shared on Hugging Face with the same code in one run: our model (\ColabOursRone\% there) is ahead of \NKkSpecificBelow{} of them, including the two E5-base models of similar size (\KkEfiveRagRone\% and \KazembedVfiveRone\%), and behind kaz-embedder (\KazEmbRone\%, three towers with \KazEmbParamsM{}M parameters in total) and a BGE-M3 fine-tune (\DarmmMultiRone\%); at recall@10 its interval overlaps those of its nearest neighbours on both sides.
Our four-language model, trained with a capped share of the same Kazakh pairs, is \KkVsCaBelKkRoneAbs{} points lower at recall@1, with overlapping recall@10 intervals.
The Kazakh model also scores higher than the released model on the Uzbek and Kyrgyz versions of Belebele, but our Tajik model, trained without any Turkic pairs, gains as much there, so these gains are not specific to Kazakh training.
Our Kazakh Wikipedia test questions come from the same generator family as the training questions, so we rest the main claims on Belebele.

\end{abstract}

\section{Introduction}

Kazakh is a Turkic language written in Cyrillic with nine letters beyond the Russian alphabet.
It has public retrieval resources, among them a question answering and retrieval set \citep{yeshpanov2024kazqad} and a retrieval benchmark \citep{tleubayeva2026kazakhtextduplicates}, yet the small open embedder EmbeddingGemma-300M \citep{vera2025embeddinggemma}, used as released, finds the right Belebele passage \citep{bandarkar2024belebele} first for \BaseBelKkRone\% of Kazakh questions.

Our earlier reports adapted language models to Tajik \citep{saidzoda2026lfm,saidzoda2026qwen3} and fine-tuned multilingual-e5-large-instruct for Tajik retrieval \citep{saidzoda2026e5}; companion reports fine-tune EmbeddingGemma-300M for Tajik \citep{saidzoda2026gemmatg} and for four Central Asian languages together \citep{saidzoda2026cagemma}.
This report covers the Kazakh model of that series: the released EmbeddingGemma-300M trained on Kazakh alone.
We ask how far one epoch of generated question and passage pairs takes it, how it compares with the published scores of other embedders and with our four-language model, and what it does to the other languages we test.

Our contributions are:
\begin{enumerate}
\item \textbf{The model and its recipe.} EmbeddingGemma-300M fine-tuned on Kazakh questions grounded in source passages, with the data pipeline and its contamination guards (Sections~\ref{sec:data} and~\ref{sec:recipe}).
\item \textbf{Position against published models.} The model gains \KkVsBaseBelKkRoneAbs{} points of Belebele recall@1; of the \NMtebModels{} models with a published MTEB score on the same test, \NMtebAbove{} score higher, at least \NMtebAboveSmall{} of them with up to \SmallCutB{}B parameters (Section~\ref{sec:results}).
\item \textbf{Kazakh-specific embedders, scored by us.} \NKkSpecific{} embedders shared on Hugging Face, scored with the same code in one run: ours is ahead of \NKkSpecificBelow{} of them and behind two larger ones (Section~\ref{sec:results}).
\item \textbf{Effects outside Kazakh.} Gains on the Uzbek and Kyrgyz versions of Belebele that a Tajik-only model also shows, and a rise in Tajik tax retrieval on a small set, with Tajik law within its interval (Section~\ref{sec:transfer}).
\end{enumerate}

\section{Four Languages, One Culture}
\label{sec:languages}

Tajik is Persian: Tajik, Iranian and Afghan Persian are mutually intelligible varieties of the same language, and Tajik is written in a modified Cyrillic alphabet \citep{merchant2025connecting}.
Kazakh, Uzbek and Kyrgyz are Turkic languages \citep{glottolog}.
The two groups share one culture, which historians call Turko-Persian \citep{green2019persianate}.
Under the Samanids, who ruled from Bukhara from the ninth century, Persian became a written language of state, and by the tenth century the court poetry of Bukhara had made it a language of government and learning in a region that later stretched from China to the Balkans.
The Turkic dynasties that ruled after the Samanids, from the Ghaznavids and Saljuqs to the Timurids and Shaybanids, kept it as their language of state.
Across much of the region Persian was ``a literate island in a sea of spoken Turkic'', the written language of courts and scholars while many nomadic peoples spoke Turkic; as \citet{green2019persianate} notes, a learned language of this kind is ``geographically broad but socially shallow''.
For centuries Bukhara, Samarkand, Tashkent and Kokand were also centres of Islamic learning, and their madrasas were still numerous on the eve of the Russian invasion: even at the end of the nineteenth century Samarkand province alone had fifty-eight, and the madrasas of Kokand attracted students even from Bukhara \citep{khalid1998politics}.
Writers of the time counted between 200 and 400 madrasas in the Emirate of Bukhara, whose students came from Iran, Afghanistan, India, Kashgar, Crimea, Tatarstan and the North Caucasus \citep{bobojonova2022madrasahs}; Bukhara was a common destination for Tatar scholars, and nomadic families hired teachers from Bukhara for their children \citep{khalid1998politics}.
Bedil, whose Persian poetry was widely read in Afghanistan and Transoxiana, was born in India to a family of the Arlas, a Chaghatay Turkic tribe \citep{ferhadi1992bidil}.

This long contact left a shared vocabulary.
The names of the days of the week in Uzbek, Kazakh and Kyrgyz come from Persian \citep{rosenfeld1996week}; Uzbek surnames are mostly of Arabic or Persian origin \citep{koziyeva2022anthroponyms}; and the Tajik of Bukhara has taken Uzbek words, suffixes and even the question particle \textit{mi}, while almost all of its speakers also speak Uzbek \citep{ido2014bukharan}.
The Russian Empire invaded the region in the 1860s and 1870s.
Under Soviet rule the madrasas were closed: by 1927 madrasa education had been abolished in Tashkent and cut sharply in Bukhara and Samarkand, before the assault on Islam of the 1930s.
The national-territorial delimitation of 1924 created national republics in place of Turkestan, Bukhara and Khiva, and the alphabet was changed twice, Uzbek for example moving from the Arabic script to Latin and, in 1940, to Cyrillic \citep{khalid2015making}.
The Soviet period also added the Russian endings \textit{-ov} and \textit{-ova} to these surnames, a custom that continues today \citep{koziyeva2022anthroponyms}.

\section{Related Work}
\label{sec:related}

\paragraph{Multilingual embedders.}
Many open multilingual embedders build on XLM-RoBERTa \citep{conneau2020unsupervised}, among them multilingual E5 \citep{wang2024me5} and BGE-M3 \citep{chen2024bgem3}; EmbeddingGemma \citep{vera2025embeddinggemma} is a 300M encoder initialised from Gemma 3 \citep{gemmateam2025gemma3}.
MTEB and MMTEB \citep{muennighoff2023mteb,enevoldsen2025mmteb} evaluate them across many languages, and their Belebele retrieval task covers Kazakh.

\paragraph{Kazakh retrieval.}
Turkic Central Asian languages have gained resources in recent years \citep{veitsman2025recent}.
KazQAD \citep{yeshpanov2024kazqad} is a Kazakh open-domain question answering dataset with a retrieval part, and KazakhTextDuplicates \citep{tleubayeva2026kazakhtextduplicates} is a peer-reviewed benchmark for duplicate detection, semantic similarity and retrieval.
A further Kazakh retrieval benchmark, a single-author preprint that is not peer-reviewed, compares morphological and dense methods on about three hundred queries \citep{seidalin2026morphology}.
In a search of arXiv and the ACL Anthology in November 2026 we found no report of a fine-tuned EmbeddingGemma for Kazakh.
Several Kazakh embedders are shared on Hugging Face or GitHub without a paper; our search was by name and is not exhaustive.
kaz-embedder\footnote{\url{https://huggingface.co/BekbolatShanakbay/kaz-embedder}} combines BGE-M3 and multilingual-e5-large-instruct towers and reports results on its own KazMTEB set; kazembed-v5\footnote{\url{https://huggingface.co/Nurlykhan/kazembed-v5}} and kazakh-e5-rag-embedding\footnote{\url{https://huggingface.co/shyngys879/kazakh-e5-rag-embedding}} fine-tune multilingual-e5-base on KazQAD and report KazQAD subsets; Darmm fine-tunes BGE-M3 and reports its own question sets.\footnote{\url{https://huggingface.co/Darmm/darmm-embedding-multilingual}, \url{https://huggingface.co/Darmm/darmm-embed-kazakh-v2}}
A comparison of Kazakh fine-tunes with BGE-M3 on the full KazQAD corpus found no fine-tune clearly ahead.\footnote{\url{https://github.com/ProgrammerXX1/kaz-embed}}
None of these reports a Belebele score, and each uses its own test, so we scored them ourselves on our Kazakh Belebele set (Section~\ref{sec:results}).
The only EmbeddingGemma variants for Kazakh we found there are copies whose vocabulary was trimmed to Kazakh text, and their cards report no retrieval evaluation.\footnote{\url{https://huggingface.co/alphaedge-ai/embeddinggemma-kaz-32768}}

\paragraph{Training on generated questions.}
Training retrievers on questions written by a language model goes back to InPars, Promptagator and GPL \citep{bonifacio2022inpars,dai2022promptagator,wang2022gpl}, and has been used for many languages at once \citep{thakur2024swimir}.
Test sets built the same way may favour systems close to the generator \citep{rahmani2024synthetic,rahmani2025bias}, which is why we read our generated test set as an upper bound.
We train with the in-batch multiple-negatives loss \citep{henderson2017smartreply} with gradient caching \citep{gao2021gradcache} in sentence-transformers \citep{reimers2019sbert}, with hard negatives mined as in NV-Retriever \citep{moreira2024nvretriever}.

\section{Data and Benchmark}
\label{sec:data}

\paragraph{Grounded question and passage pairs.}
According to our data-generation records, our in-house Central Asian fine-tune of a 27B Qwen model wrote questions about Kazakh source passages from encyclopedias, Wikipedia, news and books, and a second model, the released Qwen3.6-27B, checked that each question is answered by its passage and that the answer is correct; only questions that passed were kept, and the passage itself, not the generated answer, is the positive.
The two models are of the same family, so the check is not fully independent. The records give a pass rate of about \KkPassPct\% for the first production window only; the prompts and the overall rejection rate are not reported here, and the teacher is not released, so the generation step cannot be reproduced exactly by others.
We call these questions grounded; they are generated, not written by people.
This gave \NGroundedKk{} checked rows.
Before any pair was built, Kazakh Wikipedia articles for testing were frozen in an earlier stage of this work, as a list of article identifiers in the same form as the passage identifiers of the generated rows; we dropped \NDropHeldKk{} rows about a held-out article, \NDropTestKk{} rows whose passage is in a test corpus, \NDropNoPassageKk{} rows whose passage could not be found and \NDropDupKk{} repeated questions.
Positives that share any 13-word sequence with a Belebele or replay test passage were also to be dropped (\NDropGuardKk{} found), as were questions equal to a test question; this guard catches copied text only, not paraphrases or translations.
This leaves \NPoolKk{} pairs over \NUniqPosKk{} distinct passages.

\paragraph{Training set.}
Table~\ref{tab:data} lists the \NSetKkTotal{} pairs: all Kazakh pairs and half of the English and Russian replay sets of our Tajik model (\PctKkReplay\% of the set), from Natural Questions \citep{kwiatkowski2019nq}, MIRACL \citep{zhang2023miracl} and a Russian hard-negative set (ru-HNP).
\PctKkWithNeg\% of the Kazakh pairs carry one hard negative, mined with the released BGE-M3 \citep{chen2024bgem3} from ranks 10 to 50 of the Kazakh passage pool, from a different article than the positive and with a score below 0.95 times that of the positive \citep{moreira2024nvretriever}.

\begin{table}[t]\centering\footnotesize\setlength{\tabcolsep}{4pt}
\begin{tabular}{lr}
\toprule
Part & Pairs \\
\midrule
Kazakh, with a mined negative & \NKkNeg \\
Kazakh, in-batch negatives only & \NKkNoNeg \\
\midrule
English replay & \NKkReplayEn \\
Russian replay & \NKkReplayRu \\
\midrule
Total & \NSetKkTotal \\
\bottomrule
\end{tabular}
\caption{Training pairs of our Kazakh model, as counted by the training run.}
\label{tab:data}
\end{table}

\paragraph{Test sets.}
\begin{itemize}
\item \textbf{Belebele} \citep{bandarkar2024belebele}: \NBelKkQ{} Kazakh questions, each to be matched to its own passage among \NBelKkCorpus{} distinct passages, as in the Belebele retrieval task of MMTEB \citep{enevoldsen2025mmteb}. It is the only test not produced by our generator, so our main claims rest on it. The dataset is fully parallel, so its Tajik, Uzbek and Kyrgyz versions are one test in three translations, not independent tests.
\item \textbf{Kazakh Wikipedia}: \NWikiKkQ{} generated questions on held-out articles, searched over \NWikiKkCorpus{} passages.
\item \textbf{Tajik}: the law (\NLawQ{} questions over \NLawCorpus{} chunks), tax (\NTaxQ{} over \NTaxCorpus) and Wikipedia sets of our E5 report \citep{saidzoda2026e5}, and Belebele in Tajik, Uzbek and Kyrgyz, to look at effects outside Kazakh.
\item \textbf{Russian and English Belebele} to check that the replay languages are kept.
\end{itemize}
We report recall@1 and recall@10 with 95\% bootstrap intervals for recall@10 (1,000 resamples over questions, with the passages fixed; Belebele has several questions per passage, so these intervals are somewhat too narrow); the result files hold no intervals for recall@1, and no per-question ranks, so we cannot run paired tests.
All models are scored with the same script in bfloat16, each with its own prompts: for EmbeddingGemma ``\QueryPrompt'' and ``\DocPrompt'', for the E5 models the web-search instruction of multilingual-e5-large-instruct on queries, and none for BGE-M3.

\section{Training Recipe}
\label{sec:recipe}

\paragraph{Starting point.}
We start from the released EmbeddingGemma-300M \citep{vera2025embeddinggemma}: a Gemma 3 encoder with \NLayers{} layers, a vocabulary of \Vocab{} tokens and \EmbDim-dimensional output vectors.
Its tokenizer splits a Kazakh word into \FertKk{} tokens on average, against \FertRu{} for Russian.

\paragraph{Settings.}
The recipe is inherited from our Tajik EmbeddingGemma model \citep{saidzoda2026gemmatg} and was not tuned for Kazakh: full fine-tuning with the cached multiple-negatives ranking loss (in-batch negatives plus the mined negative, scale 20), global batch 512, learning rate \TrainLr{} with linear decay and 5\% warm-up, one epoch, sequences up to 512 tokens, seed 42, fp32 with eager attention, and the model's own query and document prompts.
Each batch is drawn from one part of the training set, parts are sampled in proportion to their size, and no text appears twice in a batch.
The epoch took \StepsKk{} steps rather than the \KkPlainSteps{} that the pair count divided by 512 would give, because each of the five parts of the training set ends in its own partial batch (\KkStepsFromParts{} batches in total); it ran on one H100 GPU shared with other jobs, with a gradient-caching mini-batch of \MiniBatchKk, which does not change the global batch.
The training pairs as counted by the run are \KkSetsMatch{} to Table~\ref{tab:data}.
The run was stopped by an operator error after about 165 steps and resumed from its step-100 checkpoint with the trainer's checkpoint resume, which restores the optimiser, the learning-rate schedule and the random state; the \KkResumeSame{} logged steps that the two segments share have identical loss values, which shows the run resumed consistently, but we did not compare it with an uninterrupted run.

\begin{figure}[t]\centering
\includegraphics[width=0.9\columnwidth]{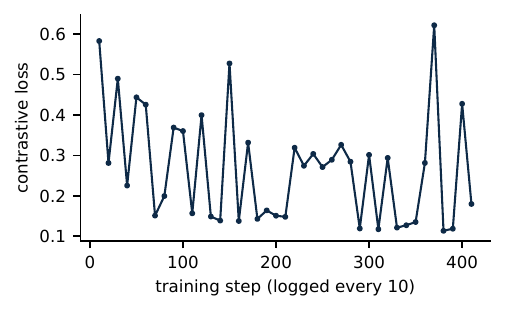}
\caption{Training loss of \ModelName, logged every 10 steps (the resumed segment overwrites the repeated steps, whose values are identical).}
\label{fig:loss}
\end{figure}

\paragraph{Training loss.}
The logged loss is that of one batch every 10 steps and varies widely from step to step (Figure~\ref{fig:loss}); it is \KkLossFirst{} at the first logged step and \KkLossLastMean{} on average over the last five, with no sign of divergence.

\section{Results}
\label{sec:results}

\begin{table*}[t]\centering\small
\begin{tabular}{lrrrrrr}
\toprule
& \multicolumn{3}{c}{Belebele Kazakh (\NBelKkQ)} & \multicolumn{3}{c}{Wikipedia Kazakh (\NWikiKkQ)} \\
Model & R@1 & R@10 & 95\% interval & R@1 & R@10 & 95\% interval \\
\midrule
\multicolumn{7}{l}{\emph{EmbeddingGemma-300M, as released and fine-tuned by us}} \\
EmbeddingGemma-300M & \BaseBelKkRone & \BaseBelKkRten & [\BaseBelKkLo, \BaseBelKkHi] & \BaseWikiKkRone & \BaseWikiKkRten & [\BaseWikiKkLo, \BaseWikiKkHi] \\
Tajik model & \TgBelKkRone & \TgBelKkRten & [\TgBelKkLo, \TgBelKkHi] & \TgWikiKkRone & \TgWikiKkRten & [\TgWikiKkLo, \TgWikiKkHi] \\
\textbf{Ours (Kazakh)} & \KkBelKkRone & \KkBelKkRten & [\KkBelKkLo, \KkBelKkHi] & \KkWikiKkRone & \KkWikiKkRten & [\KkWikiKkLo, \KkWikiKkHi] \\
Four-language model & \CaBelKkRone & \CaBelKkRten & [\CaBelKkLo, \CaBelKkHi] & \CaWikiKkRone & \CaWikiKkRten & [\CaWikiKkLo, \CaWikiKkHi] \\
\midrule
\multicolumn{7}{l}{\emph{Larger encoders (0.56B to 0.57B), scored by us}} \\
multilingual-e5-large-instruct & \EfiveBelKkRone & \EfiveBelKkRten & [\EfiveBelKkLo, \EfiveBelKkHi] & \EfiveWikiKkRone & \EfiveWikiKkRten & [\EfiveWikiKkLo, \EfiveWikiKkHi] \\
BGE-M3 & \BgeBelKkRone & \BgeBelKkRten & [\BgeBelKkLo, \BgeBelKkHi] & n/a & n/a & n/a \\
Saidzoda-Tajik-E5-Large & \EfTgBelKkRone & \EfTgBelKkRten & [\EfTgBelKkLo, \EfTgBelKkHi] & \EfTgWikiKkRone & \EfTgWikiKkRten & [\EfTgWikiKkLo, \EfTgWikiKkHi] \\
\bottomrule
\end{tabular}
\caption{Kazakh retrieval in percent. Belebele: each of the \NBelKkQ{} questions must find its own passage among \NBelKkCorpus{} Kazakh passages; the dataset is fully parallel across languages. Wikipedia: generated questions on held-out articles over \NWikiKkCorpus{} passages, written by the same generator family as our training questions, so the values are optimistic. Intervals are 95\% bootstrap intervals for recall@10. Our model is the row in bold. The Tajik model and the four-language model are our earlier EmbeddingGemma fine-tunes; Saidzoda-Tajik-E5-Large is our Tajik E5 model. Published scores of further models are in Table~\ref{tab:published}.}
\label{tab:kazakh}
\end{table*}

\paragraph{Gains over the released model.}
Table~\ref{tab:kazakh} gives the Kazakh results.
On Belebele, recall@1 rises by \KkVsBaseBelKkRoneAbs{} points, from \BaseBelKkRone{} to \KkBelKkRone, and recall@10 from \BaseBelKkRten{} to \KkBelKkRten; the recall@10 intervals \KkVsBaseBelKkOverlap.
On the generated Wikipedia set recall@1 rises from \BaseWikiKkRone{} to \KkWikiKkRone; this set starts high and its absolute values are optimistic.

\paragraph{Against larger released models.}
Used as released, the 560M XLM-RoBERTa models are ahead on Kazakh Belebele: BGE-M3 by \KkBgeGapBelKk{} points of recall@1 (\BgeBelKkRone{} against \KkBelKkRone) and multilingual-e5-large-instruct by \KkEfiveGapBelKk{} points (\EfiveBelKkRone), and in both cases the recall@10 intervals \KkVsBgeBelKkOverlap.
Our Tajik E5 model \citep{saidzoda2026e5}, which never saw Kazakh pairs, also scores higher (\EfTgBelKkRone).
On the generated Wikipedia set our model is close to multilingual-e5-large-instruct (\KkWikiKkRone{} against \EfiveWikiKkRone{} at recall@1, overlapping recall@10 intervals), but this set comes from the same generator family as our training questions and is near the ceiling at recall@10, so it says little about the gap.

To check our scoring, we compared our Belebele scores of the three released models with the published MTEB results on the same-language Kazakh subset (Table~\ref{tab:mteb}).\footnote{\url{https://github.com/embeddings-benchmark/results}, files \texttt{BelebeleRetrieval.json}, read 6 and 12 November 2026.}
They agree within \MtebMaxDiff{} points of recall@1 across the three models and all four Central Asian languages; we did not trace the remaining differences, which may come from scoring details such as precision, prompts or library versions.
This checks our scoring of the released models, not our model, which is not in MTEB.

\begin{table}[t]\centering\footnotesize\setlength{\tabcolsep}{3pt}
\begin{tabular}{lrr}
\toprule
Model & MTEB & ours \\
\midrule
EmbeddingGemma-300M & \MtebBaseBelKkRone & \BaseBelKkRone \\
mE5-large-instruct & \MtebEfiveBelKkRone & \EfiveBelKkRone \\
BGE-M3 & \MtebBgeBelKkRone & \BgeBelKkRone \\
\bottomrule
\end{tabular}
\caption{Kazakh Belebele retrieval recall@1 (percent) of the three released models, as published in the MTEB results repository and as scored by us. mE5 is multilingual-e5-large-instruct.}
\label{tab:mteb}
\end{table}

\paragraph{Against published models.}
\begin{table*}[t]\centering\small
\begin{tabular}{lrrr}
\toprule
Model & Parameters & R@1 & nDCG@10 \\
\midrule
Mira190/Euler-Legal-Embedding-V1 & 8.00B & 92.2 & 95.9 \\
Octen/Octen-Embedding-8B & 7.57B & 92.1 & 95.7 \\
Qwen/Qwen3-Embedding-8B & 7.57B & 92.0 & 95.6 \\
geevec-ai/geevec-embeddings-1.0 & API & 90.9 & 95.2 \\
google/gemini-embedding-001 & API & 90.9 & 94.7 \\
ai-sage/Giga-Embeddings-instruct-10B-A1.8B-0826 & 10.48B & 88.1 & 92.9 \\
ai-sage/Giga-Embeddings-instruct-3B-0826 & 3.15B & 87.8 & 93.0 \\
nvidia/llama-embed-nemotron-8b & 7.50B & 87.4 & 92.4 \\
tencent/KaLM-Embedding-Gemma3-12B-2511 & n/a & 87.2 & 92.3 \\
microsoft/harrier-oss-v1-27b & 27.01B & 86.9 & 92.3 \\
BAAI/bge-m3 & 0.57B & 84.0 & 90.2 \\
intfloat/multilingual-e5-large-instruct & 0.56B & 82.9 & 89.5 \\
Snowflake/snowflake-arctic-embed-l-v2.0 & 0.57B & 82.8 & 89.2 \\
jinaai/jina-embeddings-v3 & 0.57B & 82.4 & 89.1 \\
nomic-ai/nomic-embed-text-v2-moe & 0.48B & 81.7 & 88.2 \\
lightonai/mSlideOn & 0.34B & 80.7 & 87.8 \\
manu/bge-m3-custom-fr & 0.57B & 80.1 & 87.4 \\
OrdalieTech/Solon-embeddings-large-0.1 & 0.56B & 79.8 & 86.9 \\
Lajavaness/bilingual-embedding-large & 0.56B & 78.7 & 86.8 \\
\textbf{\ModelName{} (ours, scored by us)} & 0.31B & \KkBelKkRone & 84.9 \\
google/embeddinggemma-300m & 0.31B & 60.7 & 68.4 \\
openai/text-embedding-3-large & API & 47.1 & 55.6 \\
\bottomrule
\end{tabular}
\caption{Kazakh Belebele retrieval (percent): published MTEB scores (BelebeleRetrieval, kaz\_Cyrl-kaz\_Cyrl, test; results repository commit \MtebCommit, \NMtebModels{} models) and our model as scored by us; our scoring of three released models is within \MtebMaxDiff{} points of their published recall@1 over the four Central Asian languages (Kazakh in Table~\ref{tab:mteb}). Rows: the ten highest-scoring models, every model with a published count of 0.6B parameters or fewer that scores above ours, the released EmbeddingGemma-300M and OpenAI text-embedding-3-large. Parameter counts are from the MTEB model metadata; API: closed model, no count published; n/a: open model without a count in the metadata. nDCG@10 is the main MTEB score.}
\label{tab:published}
\end{table*}

Table~\ref{tab:published} places our model among all published MTEB scores on the same Kazakh Belebele subset.
Of the \NMtebModels{} models with a published score, \NMtebAbove{} score higher than ours at recall@1.
The highest, \MtebTopName, reaches \MtebTopRone, \KkTopGap{} points above ours, and the ten best are large models built on language models of several billion parameters, or closed models behind an API.
Smaller models score higher too: \NMtebAboveSmall{} open models with a published parameter count of up to \SmallCutB{}B are above ours, the smallest of them \SmallestAboveName{} with \SmallestAboveParamsM{}M parameters (\SmallestAboveRone); our model has \GemmaParamsM{}M. A further \NMtebAboveNoSize{} models above ours publish no parameter count, so the number of small models above ours is a lower bound.
So our model is not the strongest small Kazakh embedder.
Fine-tuning moves EmbeddingGemma-300M from \BaseBelKkRone{} to \KkBelKkRone, above \NMtebBelow{} of the \NMtebModels{} published models, among them OpenAI text-embedding-3-large (\OpenAiLargeRone).
The table mixes two sources, published scores and our own scoring of our model; for three released models the two differ by at most \MtebMaxDiff{} points over the four Central Asian languages (Kazakh in Table~\ref{tab:mteb}).
Kazakh-specific embedders shared by others on Hugging Face are not in MTEB, so we scored them ourselves.

\paragraph{Against Kazakh-specific embedders.}
\begin{table*}[t]\centering\small
\begin{tabular}{lrrrrr}
\toprule
Model & Parameters & R@1 & R@10 & 95\% interval & nDCG@10 \\
\midrule
BekbolatShanakbay/kaz-embedder & 1.70B & 86.0 & 96.2 & [94.8, 97.4] & 91.2 \\
Darmm/darmm-embedding-multilingual & 0.57B & 81.0 & 94.9 & [93.3, 96.3] & 88.0 \\
\textbf{\ModelName{} (ours)} & 0.31B & 77.3 & 91.4 & [89.6, 93.3] & 84.6 \\
shyngys879/kazakh-e5-rag-embedding & 0.28B & 72.6 & 91.6 & [89.7, 93.3] & 81.9 \\
Nurlykhan/kazembed-v5 & 0.28B & 69.0 & 87.9 & [85.7, 90.0] & 78.2 \\
Darmm/darmm-embed-kazakh-v2 & 0.57B & 64.3 & 85.9 & [83.6, 88.1] & 74.7 \\
alphaedge-ai/embeddinggemma-kaz-32768 & 0.13B & 57.8 & 74.3 & [71.4, 77.1] & 66.0 \\
\midrule
\multicolumn{6}{l}{\emph{Released models, scored in the same run as a check}} \\
BAAI/bge-m3 (released) & 0.57B & 83.8 & 95.8 & [94.4, 97.1] & 90.0 \\
google/embeddinggemma-300m (released) & 0.31B & 60.7 & 76.7 & [73.7, 79.6] & 68.5 \\
\bottomrule
\end{tabular}
\caption{Kazakh-specific embedders shared on Hugging Face, our model and two released models, all scored by us in one run on one \ColabGpu{} GPU with the same code and the same Kazakh Belebele set (percent; \NBelKkQ{} questions, \NBelKkCorpus{} passages). Each model is used as its card prescribes: kaz-embedder with its separate query and document towers (parameters: all three towers), the two E5-base models with query and passage prefixes, EmbeddingGemma models with their own prompts, the others without a prompt. Our three earlier-scored models reproduce their earlier recall@1 within \ColabMaxAnchorDiff{} points, so this run is comparable with Table~\ref{tab:kazakh} but its rows should be compared with each other. Intervals: 95\% bootstrap intervals for recall@10.}
\label{tab:kkspecific}
\end{table*}

Table~\ref{tab:kkspecific} gives \NKkSpecific{} Kazakh embedders shared on Hugging Face (Section~\ref{sec:related}), our model and two released models, all scored in one run with the same code on the same Kazakh Belebele set.
Our three earlier-scored models reproduce their earlier recall@1 within \ColabMaxAnchorDiff{} points (our model: \ColabOursRone{} here, \KkBelKkRone{} in Table~\ref{tab:kazakh}), so the run is comparable, and its rows are compared with each other.
Two of the Kazakh-specific models are ahead of ours, as is the released BGE-M3 (\ColabBgeRone): kaz-embedder (\KazEmbRone{} at recall@1), which encodes queries and documents with separate BGE-M3 towers and has \KazEmbParamsM{}M parameters in all its towers, with recall@10 intervals that \OursVsKazEmbOverlap{} with ours, and a BGE-M3 fine-tune by Darmm (\DarmmMultiRone, \DarmmMultiParamsM{}M parameters, overlapping recall@10 intervals).
Ours is ahead of the other \NKkSpecificBelow: the two E5-base models of similar size, kazakh-e5-rag-embedding (\KkEfiveRagRone) and kazembed-v5 (\KazembedVfiveRone), a second Darmm model (\DarmmKkVtwoRone) and the vocabulary-trimmed EmbeddingGemma (\AlphaTrimRone).
At recall@1 our lead over kazakh-e5-rag-embedding is \OursVsKkEfiveRagRoneAbs{} points, but at recall@10 the two are level (\ColabOursRten{} against \KkEfiveRagRten) with intervals that \OursVsKkEfiveRagOverlap; against kazembed-v5 the recall@10 intervals \OursVsKazembedVfiveOverlap{} as well, by a small margin.
So among the Kazakh-specific models of our size, ours is first at recall@1, and the recall@10 intervals do not separate it from the two E5-base models.

\paragraph{Against our four-language model.}
Our four-language model \citep{saidzoda2026cagemma} was trained from our Tajik model on a capped sample of the same Kazakh pairs together with Uzbek, Kyrgyz and Tajik pairs.
On Kazakh it is \KkVsCaBelKkRoneAbs{} points below the Kazakh model at Belebele recall@1 (\CaBelKkRone{} against \KkBelKkRone) with overlapping recall@10 intervals, and \KkVsCaWikiKkRoneAbs{} points above it on the generated Wikipedia set.
With one seed per model and no interval for recall@1, the Belebele difference may or may not be real.

\paragraph{Effects on the other Belebele languages.}
\label{sec:transfer}
\begin{table}[t]\centering\footnotesize\setlength{\tabcolsep}{3.5pt}
\begin{tabular}{lrrrr}
\toprule
Model & tg & kk & uz & ky \\
\midrule
Released & \BaseBelTgRone & \BaseBelKkRone & \BaseBelUzRone & \BaseBelKyRone \\
Ours (Kazakh) & \KkBelTgRone & \KkBelKkRone & \KkBelUzRone & \KkBelKyRone \\
Tajik model & \TgBelTgRone & \TgBelKkRone & \TgBelUzRone & \TgBelKyRone \\
Four-language & \CaBelTgRone & \CaBelKkRone & \CaBelUzRone & \CaBelKyRone \\
\bottomrule
\end{tabular}
\caption{Belebele recall@1 (percent) in all four languages. The Kazakh and the Tajik model were each trained on one language only, plus English and Russian replay; the four-language model saw pairs in all four.}
\label{tab:transfer}
\end{table}

\begin{figure}[t]\centering
\includegraphics[width=\columnwidth]{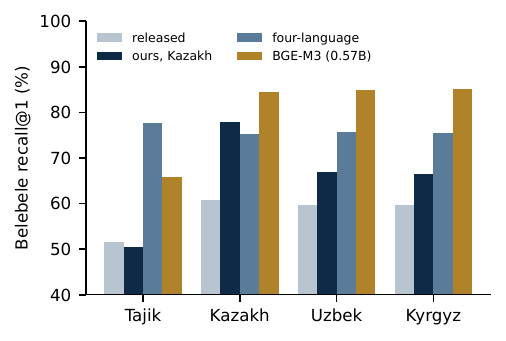}
\caption{Belebele recall@1 per language: EmbeddingGemma-300M as released, our Kazakh model, our four-language model, and BGE-M3 as released.}
\label{fig:bel}
\end{figure}

Table~\ref{tab:transfer} gives Belebele recall@1 in all four languages, and Figure~\ref{fig:bel} draws it; the four columns are translations of one question set.
Training on Kazakh alone raises Uzbek by \KkVsBaseBelUzRoneAbs{} points and Kyrgyz by \KkVsBaseBelKyRoneAbs{} points; in Uzbek the recall@10 intervals of the released and the Kazakh model \KkVsBaseBelUzOverlap, and in Kyrgyz they \KkVsBaseBelKyOverlap.
Tajik Belebele does not improve (\KkBelTgRone{} against \BaseBelTgRone).
But the Tajik model, trained without any Turkic pairs, raises Kazakh, Uzbek and Kyrgyz as well, by \TgVsBaseBelKkRoneAbs, \TgVsBaseBelUzRoneAbs{} and \TgVsBaseBelKyRoneAbs{} points, as much as the Kazakh model in Uzbek and Kyrgyz. So these gains are not specific to Kazakh training; general retrieval training or adaptation to the Belebele format would explain them as well as closeness of the languages, and we did not separate these.
In Kyrgyz, the Kazakh model (\KkBelKyRone) is close to the Tajik model (\TgBelKyRone), and far below the four-language model, which saw Kyrgyz pairs (\CaBelKyRone).

\paragraph{Tajik domain text and the replay languages.}
\begin{table}[t]\centering\footnotesize\setlength{\tabcolsep}{3.5pt}
\begin{tabular}{lrrr}
\toprule
Test set & Released & Ours & Change \\
\midrule
\multicolumn{4}{l}{\emph{Recall@10}} \\
Tajik law & \BaseLawRten & \KkLawRten & \KkVsBaseLawRten \\
Tajik tax & \BaseTaxRten & \KkTaxRten & \KkVsBaseTaxRten \\
Tajik Wikipedia & \BaseWikiRten & \KkWikiRten & \KkVsBaseWikiRten \\
Belebele Russian & \BaseRuRten & \KkRuRten & \KkVsBaseRuRten \\
Belebele English & \BaseEnRten & \KkEnRten & \KkVsBaseEnRten \\
\midrule
\multicolumn{4}{l}{\emph{Recall@1}} \\
Belebele Russian & \BaseRuRone & \KkRuRone & \KkVsBaseRuRone \\
Belebele English & \BaseEnRone & \KkEnRone & \KkVsBaseEnRone \\
\bottomrule
\end{tabular}
\caption{Tajik and the replay languages (percent), EmbeddingGemma-300M as released and our Kazakh model. Law: \NLawQ{} questions; tax: \NTaxQ.}
\label{tab:other}
\end{table}

Table~\ref{tab:other} shows the effect on our Tajik domain sets and on Russian and English.
Tajik tax recall@10 rises by \KkVsBaseTaxRtenAbs{} points on its \NTaxQ{} questions (intervals \KkVsBaseTaxOverlap, but only just: \BaseTaxHi{} and \KkTaxLo), a point estimate on a small set that we cannot explain, since Tajik Belebele does not improve; Tajik law rises by \KkVsBaseLawRtenAbs{} points (intervals \KkVsBaseLawOverlap). Both stay far below our Tajik model (\TgLawRten{} on law).
Russian and English show no measurable change: recall@10 moves by \KkVsBaseRuRten{} and \KkVsBaseEnRten{} points with overlapping intervals, and recall@1 by \KkVsBaseRuRone{} and \KkVsBaseEnRone, differences we cannot resolve without intervals.

\section{Discussion}
\label{sec:discussion}

\paragraph{When to use which model.}
For Kazakh alone, the published scores favour other models: large or closed embedders score far higher on Belebele, and at least \NMtebAboveSmall{} open models of up to \SmallCutB{}B parameters score higher as well (Table~\ref{tab:published}).
Among the Kazakh-specific models we scored, kaz-embedder is clearly ahead, with separate towers totalling \KazEmbParamsM{}M parameters, and at our size our model leads at recall@1 but is level with two E5-base models at recall@10.
Our model has fewer parameters than every model above it that publishes a parameter count, and \EmbDim-dimensional vectors, but we did not measure its speed or memory against them, so any advantage on small hardware is untested.
Where Tajik, Uzbek or Kyrgyz are also needed, our four-language model is \KkVsCaBelKkRoneAbs{} points lower on Kazakh Belebele recall@1, with overlapping recall@10 intervals, and serves all four languages.

\paragraph{What we cannot explain yet.}
Fine-tuning closes \KkVsBaseBelKkRoneAbs{} of the \BgeVsBaseBelKkRoneAbs{} points of recall@1 that separated the released EmbeddingGemma from BGE-M3 on Kazakh Belebele, but not the rest.
Our pairs are questions on encyclopedia, Wikipedia, news and book passages, while Belebele asks reading-comprehension questions about short news and travel passages; the smaller model, the different pretraining and this difference in question style are all candidates, and we did not test them.
We also did not test a second epoch, a different replay share, or training from our Tajik model instead of the released one.

\section{Release}
\label{sec:release}

\paragraph{What we release.}
The model is on the Hugging Face Hub as \texttt{Tohirju/\allowbreak\ModelName} in sentence-transformers format with its prompts, together with its training log and the result files behind every score in this report.
Access is given on request: the repository is public, and each request is approved by us.
The training pairs and the code are available on request.

\paragraph{Licence.}
EmbeddingGemma-300M is distributed under the Gemma Terms of Use (\texttt{ai.google.dev/gemma/terms}), and our weights are a model derivative under those terms: the repository states that the weights were modified, includes the Terms and the Gemma Prohibited Use Policy, and passes their use restrictions on to every user, and the access gate requires accepting them.
In addition, our model is offered under the Saidzoda model licence: organisations with revenue need a paid licence, while startups and individuals without revenue may use it free of charge until they earn from it; where the two conflict, the Gemma Terms govern.
Access requests go through \texttt{ai.saidzoda.com/access}.

\paragraph{How to use it.}
Load the model with sentence-transformers, encode questions with the ``query'' prompt and passages with the ``document'' prompt, and rank passages by cosine similarity, with inputs of up to 512 tokens.
Retrieved text should be checked before an answer reaches a user.

\section{Conclusion}

One epoch on \NPoolKk{} generated Kazakh questions grounded in source passages raised the Kazakh Belebele recall@1 of EmbeddingGemma-300M from \BaseBelKkRone{} to \KkBelKkRone.
Of \NMtebModels{} models with a published MTEB score on the same test, \NMtebAbove{} score higher, among them at least \NMtebAboveSmall{} open models of up to \SmallCutB{}B parameters, so the model is not the strongest small Kazakh embedder; our scores of three released models are within \MtebMaxDiff{} points of their published results.
Among \NKkSpecific{} Kazakh-specific embedders we scored ourselves, ours is ahead of \NKkSpecificBelow{} and behind two larger models.
Our four-language model is \KkVsCaBelKkRoneAbs{} points lower than the Kazakh model on Kazakh Belebele recall@1.
The gains on the Uzbek and Kyrgyz versions of Belebele are matched by a Tajik-only model, so they are not evidence of transfer from Kazakh.
The missing evaluation is a set of questions written by Kazakh speakers about real documents.

\section*{Limitations}

\paragraph{Generated test questions.}
The Kazakh Wikipedia questions, and the Tajik law, tax and Wikipedia questions, were written by language models of the same family as the training questions, on disjoint articles and documents, so their scores are upper bounds.
Belebele is independent, but it consists of reading-comprehension questions about short news and travel texts, not the questions people ask a search system, and it is fully parallel across languages, so the four languages in Table~\ref{tab:transfer} are one test in four translations, not independent samples. The contamination guard catches copied text only; Belebele passages come from public web sources, and paraphrased or translated overlap with our training passages is not ruled out.

\paragraph{Statistics.}
The intervals are per-model bootstrap intervals for recall@10 that resample questions only; we have no paired tests and no intervals for recall@1, and most of our comparisons are at recall@1. We treat a difference as a separation only when the recall@10 intervals do not overlap; recall@1 figures are descriptive. The Kazakh Belebele gain of \KkVsBaseBelKkRtenAbs{} points at recall@10 is far larger than the interval half-widths of \KkHalfMin{} to \KkHalfMax.
Every model was trained once, with one seed.
The tax set has only \NTaxQ{} questions.

\paragraph{Confounds.}
The four-language model starts from our Tajik model and was trained for a second epoch on a capped sample of the Kazakh pairs, so its comparison with this model mixes several differences.
The run was stopped and resumed once from a checkpoint; the repeated steps logged identical losses, but we did not compare against an uninterrupted run.
Training ran in fp32 while all scoring is in bfloat16.
The negatives were mined with the released BGE-M3.

\paragraph{Scope.}
Other models enter through their published MTEB scores, which come from several MTEB versions; we rescored only three released models, none of them a large model, and we cannot rule out that some published models saw Belebele-like data in training. We scored the Kazakh-specific embedders in a separate run on a different GPU; three earlier-scored models reproduce within \ColabMaxAnchorDiff{} points, so differences of that size between the two runs are not meaningful. Each was used as its card prescribes, and another prompt choice could change its score; kaz-embedder was scored only through its retrieval towers. We did not evaluate on KazQAD or other Kazakh benchmarks.
We tested retrieval only, with inputs of up to 512 tokens, in bfloat16 scoring.
We report no latency: the timing in our result files was measured on shared, busy cards.

\section*{Use of AI Assistants}

We used AI assistants (Anthropic Claude models) to write data-processing, training and analysis scripts, to search for and check references, and to draft parts of the text.
The training and Wikipedia test questions were generated by our own language models, as described in Section~\ref{sec:data}.
The authors designed the study, ran the training, checked every number against the recorded files, and take full responsibility for the content.
Every number in this report is generated by a script from logged files, and each reference was checked against its arXiv, ACL Anthology or DOI record.
AI assistants are not authors.

\section*{Acknowledgements}

We thank NVIDIA for access to the NVIDIA Inception programme, through which the training compute for this work was provided.

\bibliography{refs}
\bibliographystyle{acl_natbib}

\appendix
\section{Example Test Questions}
\begin{table*}[t]\centering\small
\begin{tabular}{p{0.5\textwidth}p{0.4\textwidth}}
\toprule
Question (Kazakh) & English gloss \\
\midrule
\tj{Томск облысы мен Краснояр өлкесі жер аумақтарынан өтетін Индыке өзенінің жалпы ұзындығы қанша}? & What is the total length of the Indyke river, which flows through Tomsk Oblast and Krasnoyarsk Krai? \\
\tj{Заңда адам әрекеттері дұрыс құқықты және дұрыс емес құқықты деп қалай жіктеледі}? & How does the law classify human actions into lawful and unlawful ones? \\
\tj{Сүйекқап сүйектің қалпына келуі мен өсуіне қалай қатысады}? & How does the periosteum take part in the repair and growth of bone? \\
\bottomrule
\end{tabular}
\caption{Example questions from the Kazakh Wikipedia test set, exactly as written by the generator. The English glosses are ours.}
\label{tab:examples}
\end{table*}

\end{document}